\documentclass[10pt,twocolumn,letterpaper]{article}

\usepackage[ruled,vlined]{algorithm2e}

\usepackage[pagenumbers]{cvpr} 

\usepackage{nicematrix}
\usepackage{pifont}
\newcommand{\cmark}{\ding{51}}
\newcommand{\xmark}{\ding{55}}

\usepackage{tabu}

\usepackage{xcolor}
\usepackage{colortbl}
\definecolor{SkyBlue}{RGB}{190, 230, 255}
\definecolor{Red}{RGB}{255, 0, 0}

\usepackage{musicography}

\definecolor{cvprblue}{rgb}{0.21,0.49,0.74}
\usepackage[pagebackref,breaklinks,colorlinks,citecolor=cvprblue]{hyperref}

\definecolor{commentcolor}{RGB}{110,154,155}   

\usepackage{multirow}

\def\paperID{1363} 
\def\confName{CVPR}
\def\confYear{2024}

\author{
Qin Liu$^1$~~~~~
Jianfeng Wang$^2$~~~~~
Zhengyuan Yang$^2$~~~~~
Linjie Li$^2$~~~~~ \\
Kevin Lin$^2$~~~~~ 
Marc Niethammer$^1$~~~~~
Lijuan Wang$^2$ \\
$^1$UNC Chapel Hill~~~
$^2$Microsoft \\
{\tt\small \href{https://github.com/uncbiag/LiVOS}{https://github.com/uncbiag/LiVOS}}
}

\title{LiVOS: Light Video Object Segmentation with Gated Linear Matching}

\begin{document}
\maketitle
\begin{abstract}
Semi-supervised video object segmentation (VOS) has been largely driven by space-time memory (STM) networks, which store past frame features in a spatiotemporal memory to segment the current frame via softmax attention. However, STM networks face memory limitations due to the quadratic complexity of softmax matching, restricting their applicability as video length and resolution increase. To address this, we propose \textbf{LiVOS}, a lightweight memory network that employs linear matching via linear attention, reformulating memory matching into a recurrent process that reduces the quadratic attention matrix to a constant-size, spatiotemporal-agnostic 2D state. To enhance selectivity, we introduce gated linear matching, where a data-dependent gate matrix is multiplied with the state matrix to control what information to retain or discard. Experiments on diverse benchmarks demonstrated the effectiveness of our method. It achieved \textbf{64.8} $\mathcal{J\&F}$ on MOSE and \textbf{85.1} $\mathcal{J\&F}$ on DAVIS, surpassing all non-STM methods and narrowing the gap with STM-based approaches. For longer and higher-resolution videos, it matched STM-based methods with \textbf{53\%} less GPU memory and supports \textbf{4096p} inference on a 32G consumer-grade GPU--a previously cost-prohibitive capability--opening the door for long and high-resolution video foundation models. 
\end{abstract}
\section{Introduction}
\label{sec:intro}
Video object segmentation (VOS) involves separating specific objects from the background in a sequence of video frames. It plays a critical role in various real-world applications, including robotics~\cite{petrik2022learning}, video editing~\cite{ke2021occlusion}, and medical imaging~\cite{shvets2018automatic}. Based on the level of user input, VOS tasks are generally categorized into three settings: unsupervised, semi-supervised (or one-shot), and interactive~\cite{gao2023deep}. In this work, we focus on the semi-supervised setting, where the segmentation of the first frame is provided, and the model is required to propagate this segmentation across subsequent frames. 

\begin{figure}[t]
    \centering
    \begin{subfigure}{0.48\textwidth}
        \includegraphics[width=\textwidth]{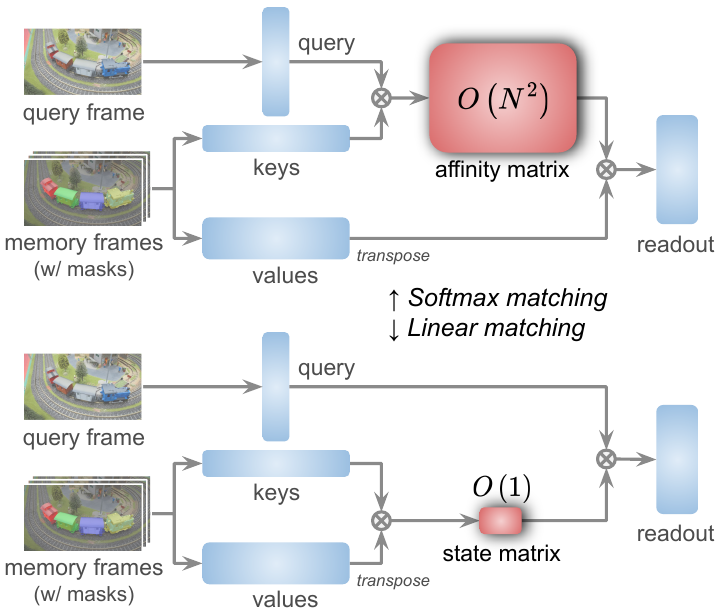}
    \end{subfigure}
    \begin{subfigure}{0.44\textwidth}
        \includegraphics[width=\textwidth]{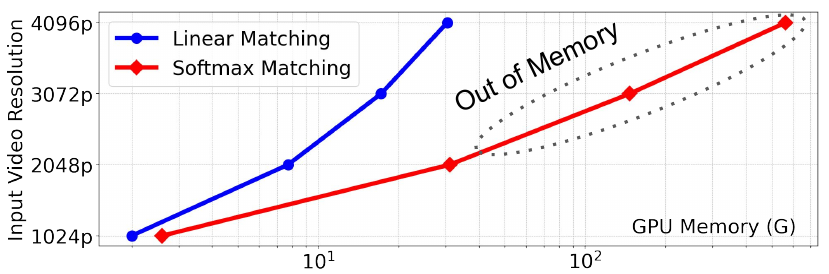}
    \end{subfigure}
    \caption{\textbf{Top}: Conceptual comparison of \emph{softmax} vs. \emph{linear} matching in video object segmentation. \textbf{Bottom}: Softmax matching suffers from memory constraints due to its \emph{quadratic} attention complexity, while linear matching achieves linear growth with a \emph{constant-size} state. Curves are based on results in Tab.~\ref{tab:quantitative_highres_videos}.}
  \label{fig:teaser}
\end{figure}

\begin{figure}
    \centering
    \begin{subfigure}[b]{0.243\textwidth}
        \centering
        \includegraphics[width=\linewidth]{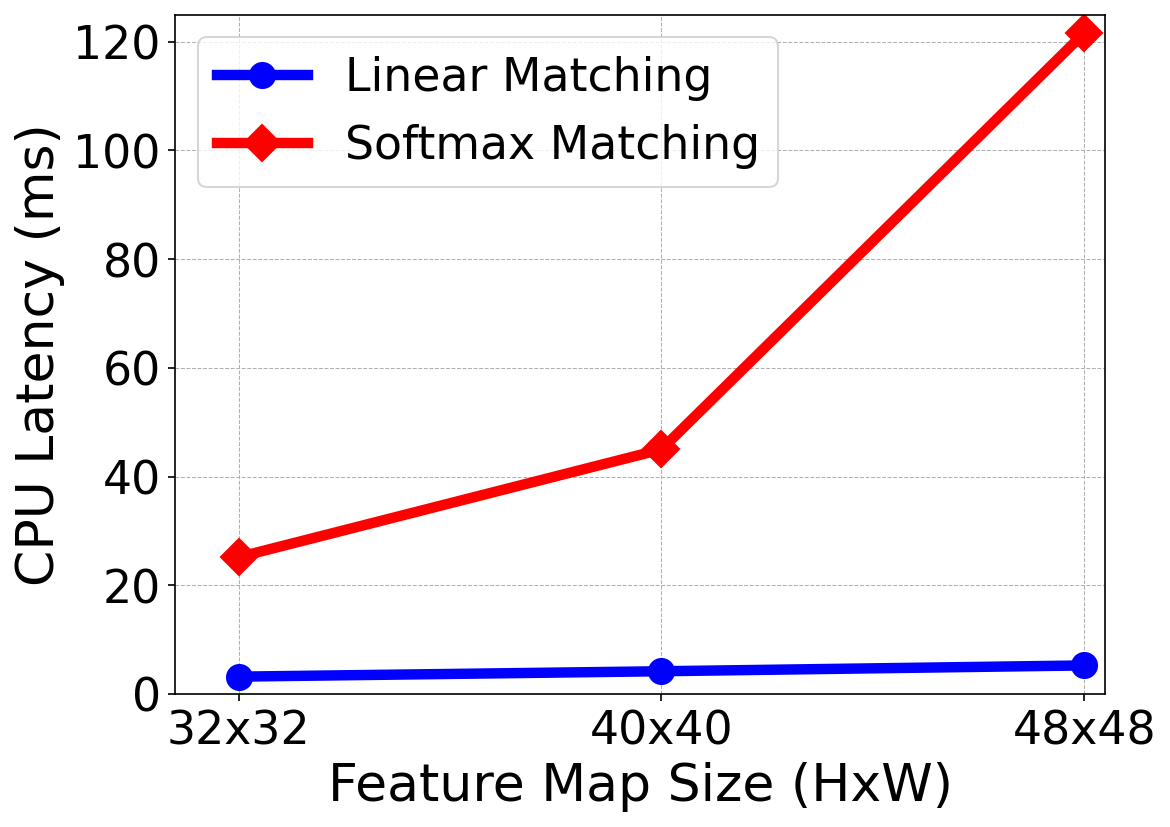}
    \end{subfigure}    
    \begin{subfigure}[b]{0.227\textwidth}
        \centering
        \includegraphics[width=\linewidth]{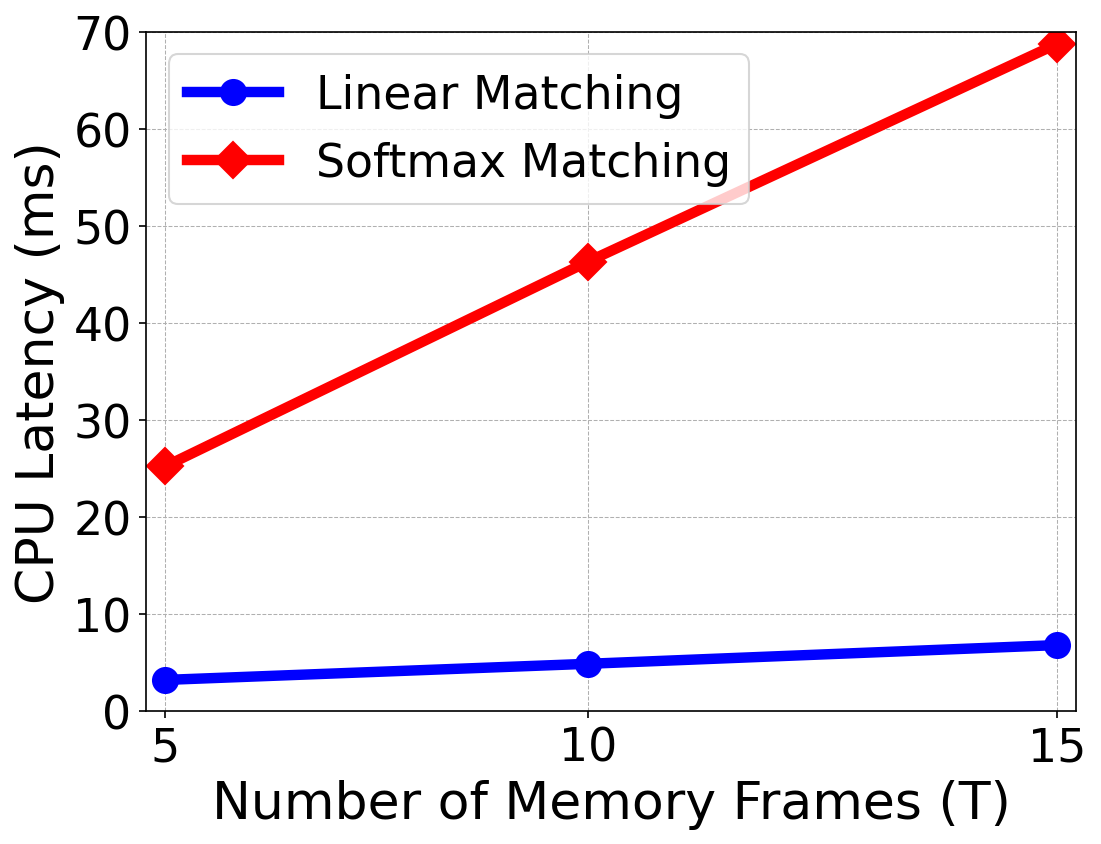}
    \end{subfigure}    
    \caption{\emph{CPU latency comparison between softmax matching and linear matching}. Softmax attention scales linearly over time (i.e., the number of memory frames $T$) and quadratically with input resolution $HW$. Latency is measured on an Intel Core-i7 (2.80GHz) CPU with PyTorch 2.0, batch size 1, and fp32.}
    \label{fig:cpu_latency}
\end{figure}

Semi-supervised VOS is primarily driven by space-time memory (STM) networks~\cite{oh2019video,cheng2021rethinking,cheng2022xmem,cheng2024putting,ravi2024sam}, which store past frames and their segmentations as \emph{keys} and \emph{values} in the memory bank to help segment the \emph{query} frame. Segmentation occurs by matching each pixel in the query frame to all spatiotemporal pixels in the memory frames via softmax attention~\cite{vaswani2017attention}. This \emph{softmax matching} process is highly inefficient due to the large attention matrix with space and time complexities of $\mathcal{O}(HW\times THW)$, where $H$ and $W$ indicate the spatial dimensions of an image and $T$ is the number of memory frames. Here the number of objects is not considered for simplicity. As shown in Fig.~\ref{fig:cpu_latency}, the CPU latency of softmax matching increases \emph{linearly} over time and \emph{quadratically} with spatial dimensions. This is problematic as video length or resolution increase, making computations too slow or causing out-of-memory issues (Fig.~\ref{fig:teaser}). 

\begin{figure}
    \centering
    \centering
    \includegraphics[width=0.96\linewidth]{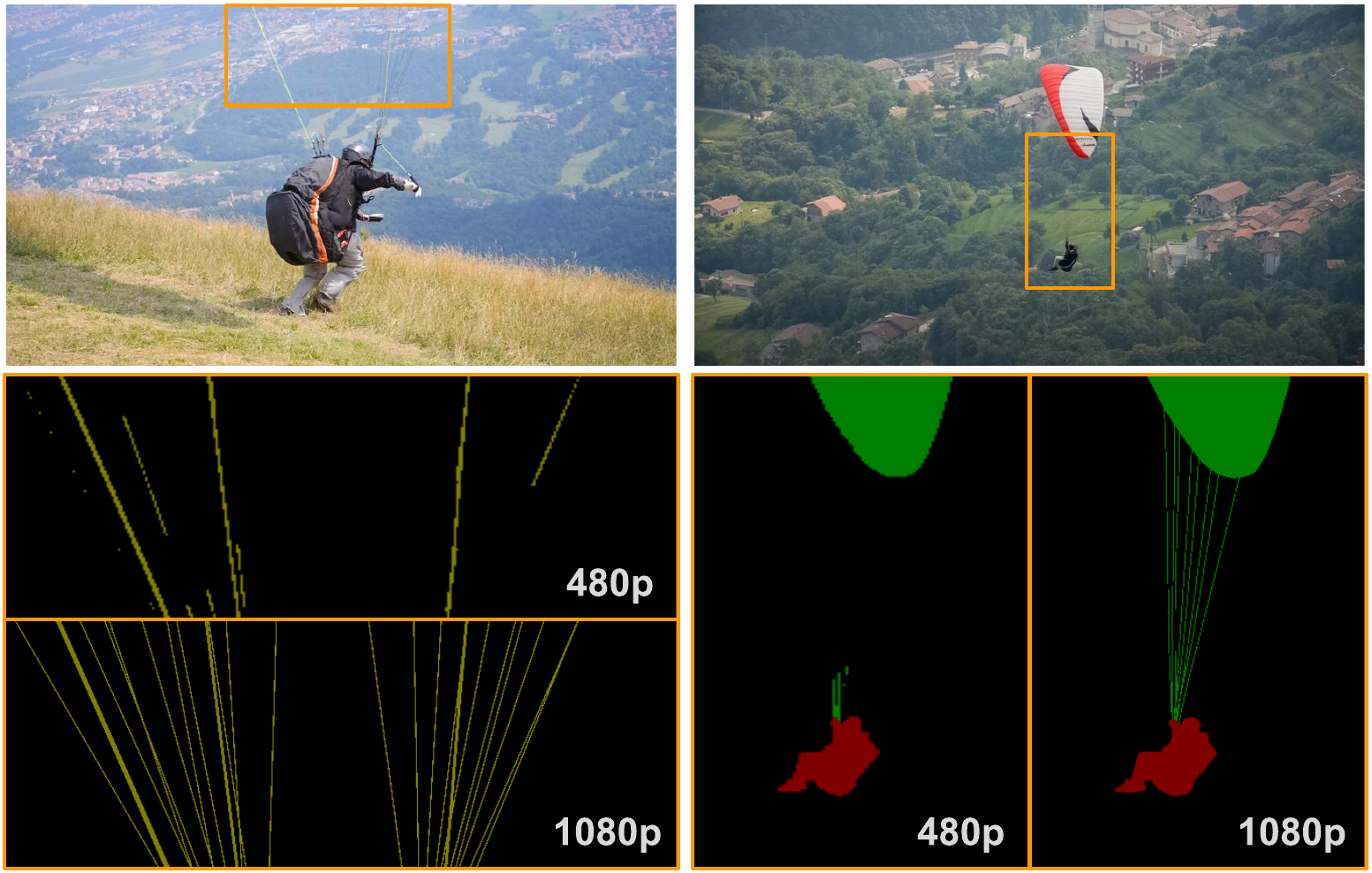}
    \caption{\emph{Masks of thin structures at different resolutions.} Thin structures may lose fine details at 480p, the standard resolution for VOS. However, existing memory networks relying on softmax matching are not efficient for processing high resolution videos.}
    \label{fig:high-resolution-masks}
\end{figure}

To alleviate these issues, one could limit the number of memory frames or downsample the video's spatial resolution. However, a fixed-size memory bank may lead to failures in challenging scenarios such as for occlusions or fast-moving objects~\cite{cheng2022xmem}. Further, downsampling a video's spatial resolution may lose fine details in object masks, as shown in Fig.~\ref{fig:high-resolution-masks}, leading to inaccurate segmentations. Therefore, memory networks relying on softmax matching face significant challenges as video lengths and resolutions increase, limiting their applicability on resource-constrained devices.

To address these challenges, strategies such as knowledge distillation~\cite{miles2023mobilevos}, recurrent feature embedding~\cite{li2022recurrent}, and spatiotemporal redundancy compressing~\cite{wang2021swiftnet} have been explored. However, these methods still rely on \emph{softmax} matching, which we identify as the \emph{core} memory bottleneck. We address this limitation by introducing \emph{linear} matching, reformulating the memory matching process into a recurrent framework that reduces the large attention matrix to a constant-size 2D state, yielding space and time complexities of $\mathcal{O}(THW)$. To enhance selectivity, we further introduce gated linear matching, where the state is multiplied by a data-dependent transition matrix~\cite{yang2023gated} that determines what information to retain or discard. Additionally, we leverage lightweight sensory memory~\cite{cheng2022xmem} and object memory~\cite{cheng2024putting} to improve performance (we claim no contribution for these components). We thus present \textbf{LiVOS}, the first light memory network that maintains \emph{constant} memory usage for arbitrarily long videos and achieves \emph{linear} memory growth as video resolution increases.

We evaluated LiVOS on a diverse range of video benchmarks, including both short- and long-term video datasets, as well as high-resolution videos. It achieved \textbf{64.8} $\mathcal{J\&F}$ on MOSE and \textbf{85.1} $\mathcal{J\&F}$ on DAVIS, surpassing all non-STM methods and narrowing the gap with STM-based approaches. For longer and higher-resolution videos, it matched STM-based methods with \textbf{53\%} less GPU memory and supports \textbf{4096p} inference on a 32G consumer-grade GPU, which is not easily possible with softmax attention. While our method is not optimized for high-resolution videos, it paves the way for developing foundation models tailored to the high-resolution regime.

In summary, we introduce a lightweight memory network that: 1) employs \emph{linear} instead of \emph{softmax} attention for memory matching, enabling efficient video object segmentation, 2) maintains constant memory usage regardless of video length, and 3) supports 4096p high-resolution inference on 32G consumer-grade hardware. Our method demonstrates competitive performance across various video benchmarks, offering computational efficiency compared to state-of-the-art methods.
\section{Related Work}
\label{sec:relatedwork}

\paragraph{Semi-supervised VOS.}
Semi-supervised VOS methods propagate sparse manual annotations, typically provided as one or more labeled frames, across the entire video sequence. Note that the term ``semi-supervised'' refers to the level of supervision required during inference, not training~\cite{perazzi2016benchmark}. Early VOS methods~\cite{badrinarayanan2010label,avinash2014seamseg,jain2014supervoxel,marki2016bilateral,perazzi2015fully} solve an optimization problem with an energy defined over a graph structure with hand-crafted energy terms. With the success of deep learning, various deep networks~\cite{tokmakov2023breaking,hu2018videomatch,caelles2017one,ventura2019rvos,wang2017learning,bhat2020learning,tokmakov2017learning,bertasius2021space,bertasius2020classifying,voigtlaender2019feelvos,wang2023look} have been proposed, including propagation-based~\cite{perazzi2017learning}, detection-based~\cite{caelles2017one}, and hybrid methods~\cite{yang2018efficient,oh2018fast}. Among these deep networks, online learning~\cite{caelles2017one,maninis2018video,voigtlaender2017online} is a common practice in which networks are trained or fine-tuned during test time. However, online learning results in slow inference, which limits its practical use. To eliminate the need for online adaptation, offline methods~\cite{oh2019video,cheng2021modular,cheng2021rethinking,cheng2022xmem,yang2022decoupling,cheng2023tracking,cheng2024putting} design deep networks that can perform object-agnostic segmentation at test time. Our method belongs to offline learning approaches; specifically, it uses a memory network with linear matching.

\paragraph{Memory-based VOS.}
Space-time memory (STM) networks~\cite{oh2019video,yang2021associating,cheng2022xmem,cheng2024putting,cheng2021rethinking} are popular memory-based methods. The seminal work of Oh \etal~\cite{oh2019video} stores past frames and their object masks as key-value pairs to help segment the current frame, treated as a query. Segmentation is performed by matching each spatial position in the query to all spatiotemporal positions in the memory frames via softmax attention~\cite{vaswani2017attention}. This softmax matching mechanism has been widely adopted in subsequent works~\cite{cheng2021rethinking,cheng2022xmem,cheng2024putting,ravi2024sam}, with improved computational efficiency~\cite{cheng2021rethinking,yang2021associating} and hierarchical memory storage~\cite{cheng2022xmem}. Recently, Cutie~\cite{cheng2024putting} achieved state-of-the-art performance by enhancing memory readout with an object transformer using cross attention. However, such methods encounter high computational costs and out-of-memory issues as video length and resolution increase~\cite{cheng2022xmem,cheng2024putting,li2022recurrent,cheng2021rethinking,cheng2023tracking}. In contrast, our method applies linear attention for memory matching, reducing the large softmax attention matrix to a constant-size 2D state. This enables constant computational costs for arbitrarily long videos and greatly improves memory efficiency as resolution scales.

\paragraph{Efficient attention.} 
Traditional softmax attention~\cite{vaswani2017attention} suffers from significant memory bottlenecks with high-resolution videos segmentation, as it requires storing an attention matrix that is \emph{quadratic} to the number of pixels in each frame. Linear attention~\cite{katharopoulos2020transformers} reduces the computational complexity of attention from \emph{quadratic} to \emph{linear} by linearizing the attention, which can be computed recurrently as a linear recurrent neural network. While linear attention is computationally more efficient, it often comes with a trade-off in model performance. Shen~\etal~\cite{shen2021efficient} improve linear attention by proposing a separate normalization of queries and keys using a softmax function before computing the inner product. Cai~\etal~\cite{cai2022efficientvit} propose multi-scale linear attention for high-resolution dense prediction tasks, achieving state-of-the-art performance. Flash Attention~\cite{dao2022flashattention} accelerates GPU training by optimizing the use of memory hierarchy, while sparse-local attention~\cite{child2019generating} and sliding window attention~\cite{jiang2023mistral} reduce global attention costs through selective focusing on parts of the spatio-temporal volume. Recently, Yang~\etal~\cite{yang2023gated} introduced gated linear attention, an improved version of linear attention with data-dependent gates, delivering competitive performance in language modeling compared to standard softmax attention and other linear-time-inference models such as RetNet~\cite{sun2023retentive} and Mamba~\cite{gu2023mamba}. Building on these advances, we investigate linear attention for semi-supervised VOS to overcome the scalability challenges of softmax-based memory networks in long, high-resolution settings.

\paragraph{Efficient VOS.}

With the rise of long and high-resolution videos, efficiency in VOS has become crucial to handle larger data volumes and computational demands while ensuring high-quality performance. Early propagation-based methods~\cite{chen2020state, yang2018efficient} are fast and memory-efficient but less accurate. Recent memory networks~\cite{oh2019video, cheng2021rethinking} improve efficiency by updating the memory bank every five frames for long videos. XMem~\cite{cheng2022xmem} enhances memory efficiency further by bounding memory bank size and introducing a training-free memory consolidation mechanism for long-term VOS, along with a sensory memory for temporal smoothness. This hierarchical memory design is adopted by Cutie~\cite{cheng2024putting}, which uses cross-attention in an object transformer to boost performance. However, Cutie’s fixed-size memory bank still faces computational limits with increasing resolution and struggles with occlusions and fast motion. Our approach overcomes these limitations by replacing softmax matching, the primary memory bottleneck, with linear matching, ensuring constant costs for long videos and linear memory growth as resolution scales.
\section{Method}
\label{sec:method}

We propose LiVOS, a light memory network for video object segmentation. The key innovation lies in replacing \emph{softmax} matching with \emph{linear} matching in standard space-time memory networks, effectively eliminating the large attention matrix that grows linearly over time and quadratically with spatial dimensions. With linear matching, LiVOS maintains a \emph{constant-size} state matrix that is updated recurrently, making it highly efficient and well-suited for long and high-resolution videos. In the following sections, we first outline the architecture of the light memory networks (Sec.~\ref{sec:lite_memory_networks}), then present gated linear matching (Sec.~\ref{sec:gated_linear_matching}), an enhanced form of linear matching at the core of our method, and conclude with implementation details (Sec.~\ref{sec:implementation_details}).

\begin{figure*}[t]
    \centering
    \includegraphics[width=\linewidth]{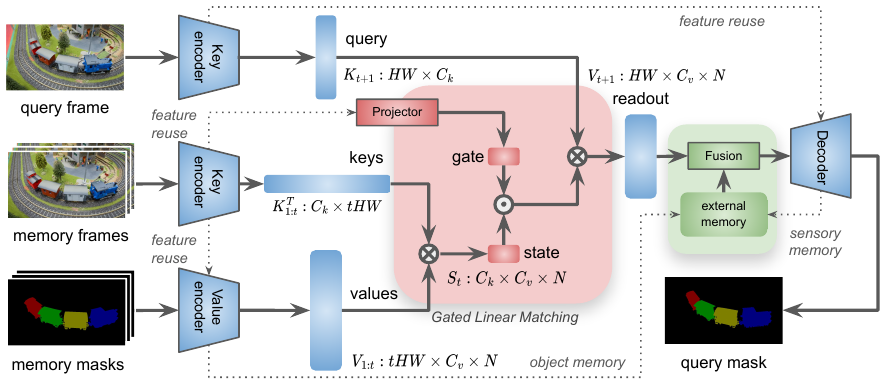}
   \caption{\emph{LiVOS Overview.} Given a query frame, we first extract its key using an image encoder and retrieve its value via gated linear matching. This value is then enhanced by two external memories before being processed by a lightweight mask decoder for segmentation. Notably, during memory matching, our method requires only a constant-size state and gate matrix. The gate matrix is generated by a gate projector that converts the pixel-level features from the last memory frame into a sparse vector, which is then transformed to a gate matrix for element-wise multiplication with the state.}
   \label{fig:method}
\end{figure*}

\subsection{Light Memory Networks}
\label{sec:lite_memory_networks}

As discussed in Sec.~\ref{sec:intro}, classical space-time memory (STM) networks are highly inefficient due to the \emph{quadratic} complexity of softmax attention, leading to memory issues as video length and resolution grow. To address this, we reduce computational complexity by using \emph{linear} attention for memory matching, replacing the large attention matrix with a \emph{lightweight}, \emph{constant-size} state matrix. This design forms the basis of our light memory network, as illustrated in Fig.~\ref{fig:method}. We first encode video frames into keys with an image encoder and masks into values with a mask encoder. We then employ gated linear matching, an enhanced form of linear matching, to generate the query frame's memory readout. Following Cutie~\cite{cheng2024putting}, we enrich the memory readout by two lightweight memories: a sensory memory~\cite{cheng2022xmem} for low-level object information and an object memory~\cite{cheng2024putting} for high-level object semantics. The sensory memory is fused with the readout via element-wise addition, while the object memory is fused via an object transformer consisting with cross-attention layers. The enriched readout is then processed by a mask decoder to generate the final segmentation mask. Additionally, we reuse features from the key encoder to enhance the features for both the value encoder and the mask decoder. In the next section, we detail gated linear matching, the core of our light memory network.

\subsection{Gated Linear Matching}
\label{sec:gated_linear_matching}

\paragraph{Notations.}
Given $t$ memory frames $\mathbf{I}_{1:t}$ and their segmentations $\mathbf{M}_{1:t}$, we extract from them memory keys $\mathbf{K}_{1:t} \in \mathbb{R}^{tHW\times C_k}$ and values $\mathbf{V}_{1:t} \in \mathbb{R}^{tHW\times C_v}$. Here, $H$ and $W$ denotes the spatial dimensions, and $C_k$ and $C_v$ are the feature dimensions for the keys and values, respectively. Given a query frame $\textbf{I}_{t+1}$, we aim to obtain its segmentation $\textbf{M}_{t+1}$. We first extract its key $\mathbf{K}_{t+1}\in \mathbb{R}^{HW\times C_k}$ and then retrieve its value $\mathbf{V}_{t+1} \in \mathbb{R}^{HW\times C_v}$ through gated linear matching. To illustrate this, we begin by introducing the traditional softmax matching.

\paragraph{Softmax Matching.} 
Softmax matching is widely used in space-time memory (STM) networks~\cite{oh2019video,cheng2021rethinking,cheng2022xmem,cheng2024putting}. Suppose we have a spatiotemporal memory consisting of memory keys $\mathbf{K}_{1:t} \in \mathbb{R}^{tHW\times C_k}$ and values $\mathbf{V}_{1:t} \in \mathbb{R}^{tHW\times C_v}$. Given a query frame $\mathbf{I}_{t+1}$, we first extract its key $\mathbf{K}_{t+1} \in \mathbb{R}^{HW\times C_k}$ and then retrieve its value $\mathbf{V}_{t+1} \in \mathbb{R}^{HW\times C_v}$ via softmax attention:
\begin{equation}
\mathbf{V}_{t+1} = Softmax(\mathbf{K}_{t+1}\mathbf{K}_{1:t}^T)\mathbf{V}_{1:t}.
\label{eq:softmax_attention_matrix}
\end{equation}

\noindent We reformulate the above softmax attention in a parallel form for $t$ memory frames:

\begin{equation}
\mathbf{V}_{t+1} = \frac{\sum_{i=1}^{t}exp(\mathbf{K}_{t+1}\mathbf{K}_{i}^T)\mathbf{V}_{i}}{\sum_{i=1}^{t}exp(\mathbf{K}_{t+1}\mathbf{K}_{i}^T)\mathbf{1}},
\label{eq:softmax_attention_exp}
\end{equation}

\noindent where $exp(\cdot)$ is an element-wise exponential for the matrix entries and $\mathbf{1} \in \mathbb{R}^{HW\times1}$ is an all-one vector. The division in the above equation is applied element-wise over the spatial dimensions $HW$. Unless specified otherwise, we will continue using this element-wise division throughout the manuscript to keep the notation concise.

Based on Eq.~\ref{eq:softmax_attention_exp}, we can write a generalized memory matching for any similarity function as follows,
\begin{equation}
\mathbf{V}_{t+1} = \frac{\sum_{i=1}^{t}sim(\mathbf{K}_{t+1}, \mathbf{K}_{i})\mathbf{V}_{i}}{\sum_{i=1}^{t}sim(\mathbf{K}_{t+1}, \mathbf{K}_{i})\mathbf{1}}.
\label{eq:softmax_attention_general}
\end{equation}

\noindent Eq.~\ref{eq:softmax_attention_general} is equivalent to Eq.~\ref{eq:softmax_attention_exp} if we use the similarity function  $sim(\mathbf{K_{t+1}},\mathbf{K_{i}})=exp(\mathbf{K_{t+1}}\mathbf{K_{i}}^T)$.

\paragraph{Linear Matching (Parallel Form).} The similarity function $sim(\cdot)$ in Eq.~\ref{eq:softmax_attention_general} needs to be non-negative. Given a kernel $k(x, y) = \langle\phi(x), \phi(y)\rangle$ with a feature representation $\phi(\cdot)$, we can rewrite Eq.~\ref{eq:softmax_attention_general} as follows,

\begin{equation}
\mathbf{V}_{t+1} = \frac{\sum_{i=1}^{t}\textcolor{red}{\phi(\mathbf{K}_{t+1})}\phi(\mathbf{K}_{i})^T\mathbf{V}_{i}}{\sum_{i=1}^{t}\textcolor{red}{\phi(\mathbf{K}_{t+1})}\phi(\mathbf{K}_{i})^T\mathbf{1}},
\label{eq:linear_matching_parallel}
\end{equation}

\noindent and then further simplify it by making use of the associative property of matrix multiplication to 

\begin{equation}
\textbf{V}_{t+1} = \frac{\textcolor{red}{\phi(\textbf{K}_{t+1})}\sum_{i=1}^{t}\phi(\textbf{K}_{i})^T\textbf{V}_{i}}{\textcolor{red}{\phi(\textbf{K}_{t+1})}\sum_{i=1}^{t}\phi(\textbf{K}_{i})^T\mathbf{1}}.
\label{eq:linear_matching_sequential}
\end{equation}

\noindent $\phi(\cdot)$ can be any non-linear function. Inspired by Shen \etal~\cite{shen2021efficient}, we set $\phi(\cdot)$ as softmax function that is applied row-wise to the keys $\mathbf{K}_{i} \in \mathbb{R}^{HW\times C_k}$.

\paragraph{Linear Matching (Recurrent Form).} 
Letting $\mathbf{S}_{t} = \sum_{i=1}^{t}\phi(\mathbf{K}_{i})^T\mathbf{V}_{i}$ and $\mathbf{Z}_t=\sum_{i=1}^{t}\phi(\mathbf{K}_{i})^T\mathbf{1}$ where $\mathbf{S}_t \in \mathbb{R}^{C_k\times C_v\times N}$ and $\mathbf{Z}_t \in \mathbb{R}^{C_k\times HW}$, we can rewrite Eq.~\ref{eq:linear_matching_sequential} in a recurrent form,

\begin{equation}
\mathbf{S}_{t} = \mathbf{S}_{t-1} + \phi(\mathbf{K}_{i})^T\mathbf{V}_{i},
\label{eq:linear_matching_state}
\end{equation}
\begin{equation}
\mathbf{Z}_{t} = \mathbf{Z}_{t-1} + \phi(\mathbf{K}_{i})^T\mathbf{1},
\label{eq:linear_matching_z}
\end{equation}
\begin{equation}
\mathbf{V}_{t+1} = \frac{\phi(\mathbf{K}_{t+1})\mathbf{S}_t}{\phi(\mathbf{K}_{t+1})\mathbf{Z}_t}.
\label{eq:linear_matching_v}
\end{equation}

\noindent Here, $\mathbf{S}_t \in \mathbb{R}^{C_k\times C_v}$ can be regarded as a 2D recurrent hidden state that is constant size. From a sequence modeling perspective~\cite{katharopoulos2020transformers}, $\phi(\mathbf{K}_{i})^T\mathbf{V}_{i}$ expands the state from the key space $\mathbb{R}^{C_k}$ and value space $\mathbb{R}^{C_v}$ into the state space $ \mathbb{R}^{C_k\times C_v}$, while $\phi(\mathbf{K}_{t+1})\mathbf{S}_t$ reduces it back. 

\paragraph{Complexity Analysis.}
For \emph{softmax} matching in Eq.~\ref{eq:softmax_attention_exp}, the space and time complexity is $\mathcal{O}(THW\times HW \times N)$, where $T$ is the number of memory frames, $H$ and $W$ are spatial dimensions, and $N$ is the number of objects. It is \emph{quadratic} in the spatial dimensions $HW$ because the attention matrix must be stored to compute the weights for values. In contrast, \emph{linear} matching has space and time complexity of $\mathcal{O}(HW\times N)$. It is \emph{linear} in the spatial dimensions $HW$ because we reduce the large attention matrix to a constant-size 2D state $S_t$. Fig.~\ref{fig:cpu_latency} shows the CPU latency comparison between linear and softmax matching. 

\paragraph{Gated Linear Matching.} 
The linear matching in Eq.~\ref{eq:linear_matching_v} does not have a selection mechanism, which has shown to be crucial in long-context tasks~\cite{buckman2024}. To enhance selectivity, we propose gated linear matching, where the state $\mathbf{S}_t \in \mathbb{R}^{C_k\times C_v}$ is multiplied element-wise by a data-dependent forget gate $\mathbf{G}_t \in (0, 1)^{C_k\times C_v}$:

\begin{equation}
\mathbf{S}_t = \mathbf{G}_t\odot \mathbf{S}_{t-1} + \phi(\mathbf{K}_{i})^T\mathbf{V}_{i},
\end{equation}

\noindent Where $\odot$ denotes element-wise multiplication. The gate $\mathbf{G}_t$ controls which information is enhanced or suppressed (more details please refer to Sec.~\ref{sec:implementation_details}). Although the concept of gated linear matching is inspired by gated linear attention~\cite{yang2023gated} in language modeling, we explore its application in the context of semi-supervised VOS.

\subsection{Fusion with External Memory}
\label{sec:fusion_with_external_memory}

After gated linear matching, we obtain the memory readout $\mathbf{V}_{t+1}$ for the query frame $\mathbf{I}_{t+1}$. Following Cutie~\cite{cheng2024putting}, we enrich it with two lightweight external memories: a sensory memory for low-level object information and an object memory for high-level object semantics. Note that we claim no contribution of the two components, and we describe these two components in the appendix.

\subsection{Implementation Details}
\label{sec:implementation_details}

\paragraph{Encoder.} We use ResNet-50~\cite{he2016deep} as the image encoder and ResNet-18~\cite{he2016deep} as the mask encoder, each producing multi-scale features: $\mathbf{f}_4$, $\mathbf{f}_8$, and the coarsest feature $\mathbf{f}_{16} \in \mathbb{R}^{HW \times C_f}$, where the subscript denotes the stride 16. Keys $\mathbf{K} \in \mathbb{R}^{HW \times C_k}$ are extract from the image encoder's coarsest feature using convolutional layers, and values $\mathbf{V} \in \mathbb{R}^{HW \times C_v}$ are similarly extracted from the mask encoder’s coarsest feature. Following Cutie~\cite{cheng2024putting}, the image encoder’s multi-scale features are reused in the mask encoder for value extraction, with $C_k = 64$ and $C_v = 256$. 

\paragraph{Decoder.} We adopt the lightweight mask decoder from Cutie~\cite{cheng2024putting} for simplicity and efficiency. This decoder takes as input the value readout $\mathbf{V}$ at stride 16 and skip connections $\mathbf{f}_4$ and $\mathbf{f}_8$ from the image encoder at strides 4 and 8, respectively. The value readout is processed with two upsampling blocks, incorporating skip connections to retain high-frequency details. Each block bilinearly upsamples the input feature by a factor of two, then adds the result to the skip-connection features. For multi-object cases, we apply soft-aggregation~\cite{oh2019video} to merge object logits.

\paragraph{Gate.} The state gate can be parameterized in various ways~\cite{yang2023gated}, depending on a balance between parameter efficiency, training efficiency, and state size. In our work, we obtain the data-dependent gate via low-rank parameterization $\mathbf{G_t} \in \mathbb{R}^{C_k\times C_v} = \mathbf{\alpha_t}\mathbf{1}^T$, where $\mathbf{1} \in \mathbb{R}^{C_v\times1}$ is an all-one vector and $\mathbf{\alpha_t} \in (0, 1)^{C_k\times 1}$ is extracted from the image encoder's coarsest feature map $\mathbf{f}_{t} \in \mathbb{R}^{HW\times C_{f}}$. To implement this, we first use a depth-wise convolutional layer to convert the feature $\mathbf{f}_{t}$ from $\mathbb{R}^{HW\times C_f}$ to $\mathbb{R}^{HW\times C_k}$, and then sum over its spatial dimensions to obtain $\hat{\mathbf{\alpha_t}} \in \mathbb{R}^{C_k\times 1}$, followed by a Sigmoid function to obtain $\mathbf{\alpha_t} \in (0, 1)^{C_k\times 1}$.

\paragraph{Training.} We use PyTorch~\cite{paszke2019pytorch} and AdamW~\cite{loshchilov2017decoupled} optimizer with an initial learning rate of $1e^{-4}$, a batch size of 16, and a weight decay of 0.001. Each training batch contains 8 frames randomly selected from a video and cropped to a default size $480\times480$. Our training lasts for 125K iterations, reducing the learning rate by a factor of 10 after 100K and 115K iterations. The state and the gate are updated for each frame in a training batch. To mitigate overfitting, we apply a learning rate multiplier $r=0.1$ to the image encoder. We also clip the global gradient norm to $\tau=3$ and use stable data augmentation~\cite{cheng2023tracking}. We use a combined loss function of cross-entropy and soft dice loss with equal weighting following~\cite{cheng2024putting,cheng2022xmem,cheng2023tracking}. Following Cutie~\cite{cheng2024putting}, we adopt point supervision with $K=12544$ sampled points instead of the whole mask for efficient training. Training takes approximately 90 hours on four NVIDIA A6000 GPUs.

\paragraph{Inference.} Unless stated otherwise, we resize the input videos so that the shorter edge is no more than 480 pixels, then rescale the model’s prediction to the original resolution. We segment video frames sequentially and update the state as well as the gate for each frame. For datasets such as YouTube VOS where new objects may appear in intermediate frames, we create a new state for each new object with negligible computational costs. Note that training has no such issue because the number of classes is fixed in the training batch. Evaluation were conducted using an NVIDIA A6000 GPU.
\section{Experiments}
\label{sec:experiments}

\subsection{Benchmarks}

\paragraph{Pretraining.} Our method relies solely on ImageNet~\cite{deng2009imagenet} pretraining, without additional pretraining on static images or synthetic videos like BL30K~\cite{miles2023mobilevos}. While static image pretraining is standard in baseline methods (Sec.~\ref{sec:baselines}), we omit it, as it showed no improvement.

\paragraph{Datasets.} We train and evaluate our method on the following datasets: 1) \textbf{DAVIS 2017}~\cite{pont20172017,perazzi2016benchmark}, a widely used VOS dataset with three main sets: 1) a training set with 60 videos (4,219 frames, 138 objects), 2) a validation set with 30 videos (2,023 frames, 59 objects), and 3) a test set with 30 videos (2,037 frames, 89 objects). 2) \textbf{YouTube-VOS 2019}~\cite{xu2018youtube}, one of the largest VOS dataset, featuring a wide range of challenging real-world scenarios. The dataset consists of a training set with 3,471 videos (197,292 frames) and a validation set with 507 videos (19,981 frames). There are 65 seen object categories in the training set and 26 unseen categories only appear in the validation set, challenging a model's generalization ability. 3) \textbf{MOSE}~\cite{ding2023mose}, a large-scale video object segmentation (VOS) benchmark emphasizing multi-object scenarios in diverse environments. It comprises 2,149 videos, divided into 1,507 training videos, 311 validation videos, and 331 testing videos. The dataset includes 5,200 objects across 36 distinct categories. In this work, we utilize the training set for model training and evaluate performance on the validation set. 4) \textbf{LVOS}~\cite{hong2022lvos}, a dataset designed specifically for video object segmentation in long, continuous video sequences, focusing on more extended temporal challenges. It's divided into three main sets: 1) a training set with 120 videos, 2) a validation set with 50 videos, and 3) a test set with 50 videos. There are 126,280 frames and 156,432 annotations in total. In this work, we report evaluation results on the validation and test sets without training our model on LVOS.

\paragraph{Evaluation Metrics.}
We report Region Jaccard $\mathcal{J}$~\cite{perazzi2016benchmark}, contour accuracy $\mathcal{F}$~\cite{fernandez2018new}, and their average $\mathcal{J\&F}$ to measure the segmentation quality. For YouTube-VOS, we additionally report $\mathcal{G}$ as the average of $\mathcal{J\&F}$ across both seen and unseen classes. We also report frames per second (FPS) and GPU memory consumption (\ie maximum GPU memory allocated by PyTorch~\cite{paszke2019pytorch}) to measure speed and resource usage, respectively. Both metrics are recorded on the same machine under similar conditions for all methods.

\subsection{Baselines}
\label{sec:baselines}

We compare LiVOS with various types of approaches, including state-of-the-art space-time memory (STM) networks, represented by Cutie~\cite{cheng2024putting} and methods without using STM, represented by RDE~\cite{li2022recurrent}. This section only introduces several representative methods. A complete introduction will be included in the appendix.

\noindent \textbf{RDE}~\cite{li2022recurrent} introduces a recurrent dynamic embedding to maintain a constant-size memory, with image and mask encoders based on ResNet-50. We use the version without BL30K~\cite{cheng2021modular} pretraining as a baseline.

\noindent \textbf{STCN}~\cite{cheng2021rethinking} is a state-of-the-art memory network building on STM~\cite{oh2019video}, enhancing robustness and efficiency. For fair comparison, we use the model without BL30K~\cite{cheng2021modular} pretraining.

\noindent \textbf{XMem}~\cite{cheng2022xmem} targets long-term videos with a compact long-term memory, a fast-updating sensory memory, and an STCN-based working memory. As with STCN, we use the model without BL30K\cite{cheng2021modular} pretraining.

\noindent \textbf{Cutie}~\cite{cheng2024putting} enhances XMem\cite{cheng2022xmem} with a top-down, object-level memory reading mechanism for improved video segmentation. We use its small and base models as baselines. We also limit Cutie to using a single memory frame during inference for fair comparison with non-STM methods.

\begin{table*}
\footnotesize
\centering
\begin{tabular}{l@{\hspace{1pt}}c c c c c c c c c c c c c c c}
    \toprule
    \multirow{2}{*}{\textbf{Method}}
    & \multirow{2}{*}{\textbf{with}}
    & \multicolumn{3}{c}{\textbf{MOSE}} 
    & \multicolumn{3}{c}{\textbf{DAVIS-17 val}} 
    & \multicolumn{3}{c}{\textbf{DAVIS-17 test}} 
    & \multicolumn{5}{c}{\textbf{YouTubeVOS-2019 val}} \\
    \cmidrule(lr){3-5} \cmidrule(lr){6-8} \cmidrule(lr){9-11} \cmidrule(lr){12-16}
    & \textbf{STM}
    & $\mathcal{J\&F}$ & $\mathcal{J}$ &  $\mathcal{F}$ 
    & $\mathcal{J\&F}$ & $\mathcal{J}$ &  $\mathcal{F}$ 
    & $\mathcal{J\&F}$ & $\mathcal{J}$ &  $\mathcal{F}$
    & $\mathcal{G}$ & $\mathcal{J}_s$ & $\mathcal{F}_s$ & $\mathcal{J}_u$ & $\mathcal{F}_u$ \\
    \midrule
    {\musQuarter} STCN~\cite{cheng2021rethinking} \tiny{NeurIPS'21}
        & \cmark
        & 52.5 & 48.5 & 56.6 & 85.4 & 82.2 & 88.6 & 76.1 & 72.7 & 79.6 
        & 82.7 & 81.1 & 85.4 & 78.2 & 85.9 \\
    {\musQuarter} AOT~\cite{yang2021associating} \tiny{NeurIPS'21}
        & \cmark
        & 58.4 & 54.3 & 62.6 & 84.9 & 82.3 & 87.5 & 79.6 & 75.9 & 83.3 
        & 85.3 & 83.9 & 88.8 & 79.9 & 88.5 \\
    {\musQuarter} XMem~\cite{cheng2022xmem} \tiny{ECCV'22}
        & \cmark
        & 56.3 & 52.1 & 60.6 & 86.2 & 82.9 & 89.5 & 81.0 & 77.4 & 84.5 
        & 85.5 & 84.3 & 88.6 & 80.3 & 88.6 \\
    {\musQuarter} DeAOT~\cite{yang2022decoupling} \tiny{NeurIPS'22}
        & \cmark
        & 59.0 & 54.6 & 63.4 & 85.2 & 82.2 & 88.2 & 80.7 & 76.9 & 84.5 
        & 85.6 & 84.2 & 89.2 & 80.2 & 88.8 \\
    {\musQuarter} DEVA~\cite{cheng2023tracking} \tiny{ICCV'23}
        & \cmark
        & 60.0 & 55.8 & 64.3 & 86.8 & 83.6 & 90.0 & 82.3 & 78.7 & 85.9 
        & 85.5 & 85.0 & 89.4 & 79.7 & 88.0 \\
    {\musQuarter} Cutie-small~\cite{cheng2024putting} \tiny{CVPR'24}
        & \cmark
        & 62.2 & 58.2 & 66.2 
        & 87.2 & 84.3 & 90.1 
        & 84.1 & 80.5 & 87.6
        & 86.2 & 85.3 & 89.6 & 80.9 & 89.0 \\
    {\musHalf} Cutie-small~\cite{cheng2024putting} \tiny{CVPR'24}
        & \cmark
        & 67.4 & 63.1 & 71.7
        & 86.5 & 83.5 & 89.5
        & 83.8 & 80.2 & 87.5 
        & 86.3 & 85.2 & 89.7 & 81.1 & 89.2 \\
    {\musQuarter} Cutie-base~\cite{cheng2024putting} \tiny{CVPR'24}
        & \cmark
        & 64.0 & 60.0 & 67.9 
        & 88.8 & 85.4 & 92.3 
        & 84.2 & 80.6 & 87.7 
        & 86.1 & 85.5 & 90.0 & 80.6 & 88.3 \\
    {\musHalf} Cutie-base~\cite{cheng2024putting} \tiny{CVPR'24}
        & \cmark
        & 68.3 & 64.2 & 72.3 
        & 88.8 & 85.6 & 91.9
        & 85.3 & 81.4 & 89.3 
        & 86.5 & 85.4 & 90.0 & 81.3 & 89.3 \\
    \midrule
    {\musQuarter} CFBI~\cite{yang2020collaborative} \tiny{ECCV'20}
        & \xmark
        & - & - & -
        & 81.9 & 79.1 & 84.6
        & 74.8 & 71.1 & 78.5 
        & - & - & - & - & - \\
    {\musQuarter} CFBI+~\cite{yang2021collaborative} \tiny{TPAMI'21}
        & \xmark
        & - & - & - 
        & 82.9 & 80.1 & 85.7
        & 75.6 & 71.6 & 79.6 
        & - & - & - & - & - \\
    {\musQuarter} SwiftNet~\cite{wang2021swiftnet} \tiny{CVPR'21}
        & \xmark
        & - & - & -
        & 81.1 & 78.3 & 83.9
        & - & - & -
        & 77.8 & 77.8 & 81.8 & 72.3 & 79.5 \\
    {\musQuarter} RDE~\cite{li2022recurrent} \tiny{CVPR'22}
        & \xmark
        & 46.8 & 42.4 & 51.3 
        & 84.2 & 80.8 & 87.5 
        & 77.4 & 73.6 & 81.2 
        & 81.9 & 81.1 & 85.5 & 76.2 & 84.8 \\
    {\musQuarter} MobileVOS~\cite{miles2023mobilevos} \tiny{CVPR'23}
        & \xmark
        & - & - & -
        & 83.7 & 80.2 & 87.1 
        & - & - & -
        & 82.3 & 81.6 & 86.0 & \textbf{76.3} & \textbf{85.2} \\
    {\musQuarter} Cutie-small$^\dag$~\cite{cheng2024putting} \tiny{CVPR'24}
        & \xmark
        & 49.3 & 45.4 & 53.2
        & 76.4 & 73.0 & 79.8 
        & 71.6 & 67.9 & 75.3 
        & 79.0 & 78.0 & 81.9 & 74.5 & 81.8 \\
    {\musHalf} Cutie-small$^\dag$~\cite{cheng2024putting} \tiny{CVPR'24}
        & \xmark
        & 51.7 & 47.7 & 55.7 
        & 74.9 & 71.7 & 78.2
        & 72.3 & 68.8 & 75.9 
        & 78.9 & 77.6 & 81.4 & 74.6 & 81.9 \\        
    {\musQuarter} Cutie-base$^\dag$~\cite{cheng2024putting} \tiny{CVPR'24}
        & \xmark
        & 50.6 & 46.6 & 54.6
        & 79.3 & 75.8 & 82.7
        & 73.5 & 70.0 & 77.0 
        & 80.1 & 79.1 & 83.3 & 75.2 & 82.7 \\
    {\musHalf} Cutie-base$^\dag$~\cite{cheng2024putting} \tiny{CVPR'24}
        & \xmark
        & 52.6 & 48.6 & 56.6
        & 77.5 & 74.1 & 80.8 
        & 73.2 & 69.4 & 76.9
        & 79.5 & 78.4 & 82.4 & 75.1 & 82.4 \\    
    \rowcolor{SkyBlue}
    {\musQuarter} LiVOS
        & \xmark
        & 59.2 & 55.6 & 62.8
        & 84.4 & 81.2 & 87.6
        & 78.2 & 74.8 & 81.7 
        & 79.9 & 82.6 & 86.8 & 71.7 & 78.4 \\
    \rowcolor{SkyBlue}
    {\musQuarter} LiVOS$^\ast$
        & \xmark
        & 58.4 & 54.7 & 62.0
        & \textbf{85.1} & \textbf{81.9} & \textbf{88.3}
        & \textbf{81.0} & \textbf{77.4} & \textbf{84.5} 
        & 81.3 & \textbf{83.6} & \textbf{87.6} & 73.3 & 80.6 \\
    \rowcolor{SkyBlue}
    {\musHalf} LiVOS
        & \xmark
        & \textbf{64.8} & \textbf{60.9} & \textbf{68.7}
        & 84.0 & 80.6 & 87.3
        & 79.6 & 76.2 & 82.9 
        & \textbf{82.6} & 83.5 & \textbf{87.6} & 75.8 & 83.3 \\
    \bottomrule
\end{tabular}
\caption{\emph{Quantitative comparisons on short videos (480p).} All methods are trained on 480p videos and use ResNet as the backbone: Cutie-small uses ResNet-18, CFBI/CFBI+ use ResNet-101, and the rest use ResNet-50. {\musQuarter} represents models trained on YouTube VOS and DAVIS, while {\musHalf} denotes additional training with MOSE. $^\dag$indicates a downgraded Cutie model that uses only one memory frame during inference. Note that completely removing the memory frames in Cutie renders the method dysfunctional. LiVOS notably improves segmentation quality over non-STM approaches and matches STM-based methods. $^\ast$training with 30k additional iterations.}
\label{tab:quantitative_short_videos}
\end{table*}
\begin{table*}
\centering
\footnotesize
\begin{tabular}{l@{\hspace{6pt}}c c c c c c c c c c c c c}
    \toprule
    \multirow{2}{*}{\textbf{Method}}
    & \multirow{2}{*}{\textbf{Key}}
    & \multirow{2}{*}{\textbf{Value}}    
    & \multirow{2}{*}{\textbf{with}} 
    & \multicolumn{5}{c}{\textbf{LVOS val}} 
    & \multicolumn{5}{c}{\textbf{LVOS test}} \\
    \cmidrule(lr){5-9}
    \cmidrule(lr){10-14}
    & \textbf{Encoder}
    & \textbf{Encoder}
    & \textbf{STM}
    & $\mathcal{J\&F}$ & $\mathcal{J}$ & $\mathcal{F}$ 
    & Mem$\downarrow$ & FPS$\uparrow$
    & $\mathcal{J\&F}$ & $\mathcal{J}$ & $\mathcal{F}$ 
    & Mem$\downarrow$ & FPS$\uparrow$ \\
    \midrule
    {\musQuarter} DEVA~\cite{cheng2024putting} \tiny{ICCV'23}
    & RN-50 & RN-18 & \cmark
    & 58.3 & 52.8 & 63.8 & 1147M & 48.3 & 54.0 & 49.0 & 59.0 & 1347M & 46.6 \\
    {\musHalf} DEVA~\cite{cheng2024putting} \tiny{ICCV'23}
    & RN-50 & RN-18 & \cmark
    & 55.9 & 51.1 & 60.7 & 1147M & 48.3 & 56.5 & 52.2 & 60.8 & 1347M & 46.6 \\
    {\musQuarter} Cutie-small~\cite{cheng2024putting} \tiny{CVPR'24}
    & RN-18 & RN-18 & \cmark
    & 58.8 & 54.6 & 62.9 & 1013M & 34.0 & 57.2 & 53.7 & 60.7 & 1094M & 32.8 \\    
    {\musHalf} Cutie-small~\cite{cheng2024putting} \tiny{CVPR'24}
    & RN-18 & RN-18 & \cmark
    & 60.7 & 55.6 & 65.8 & 1013M & 34.0 & 56.9 & 53.5 & 60.2 & 1094M & 32.8 \\
    {\musQuarter} Cutie-base~\cite{cheng2024putting} \tiny{CVPR'24}
    & RN-50 & RN-18 & \cmark
    & 60.1 & 55.9 & 64.2 & 1092M & 30.1 & 56.2 & 51.8 & 60.5 & 1175M & 29.7 \\
    {\musHalf} Cutie-base~\cite{cheng2024putting} \tiny{CVPR'24}
    & RN-50 & RN-18 & \cmark
    & 63.5 & 59.1 & 67.9 &  1092M & 30.1 & 63.6 & 59.1 & 68.0 & 1175M & 29.7 \\
    \midrule
    {\musQuarter} RDE~\cite{li2022recurrent} \tiny{CVPR'22}
    & RN-50 & RN-18 & \xmark
    & 47.2 & 41.7 & 52.7 & 9.0G$^\ddag$ & 40.6 
    & 44.7 & 39.2 & 50.2 & 12.7G$^\ddag$ & 39.8 \\
    {\musQuarter} Cutie-small$^\dag$~\cite{cheng2024putting} \tiny{CVPR'24}
    & RN-18 & RN-18 & \xmark
    & 48.2 & 43.9 & 52.5 & 585M & \textbf{56.1} 
    & 45.2 & 41.3 & 49.0 & 686M & \textbf{55.2} \\
    {\musHalf} Cutie-small$^\dag$~\cite{cheng2024putting} \tiny{CVPR'24}
    & RN-18 & RN-18 & \xmark
    & 46.1 & 42.1 & 50.1 & 585M & \textbf{56.1} 
    & 44.0 & 40.1 & 48.0 & 686M & \textbf{55.2} \\
    {\musQuarter} Cutie-base$^\dag$~\cite{cheng2024putting} \tiny{CVPR'24}
    & RN-50 & RN-18 & \xmark
    & 49.1 & 44.7 & 53.4 & 668M & 45.5 & 48.7 & 44.6 & 52.7 & 748M & 43.0 \\
    {\musHalf} Cutie-base$^\dag$~\cite{cheng2024putting} \tiny{CVPR'24}
    & RN-50 & RN-18 & \xmark
    & 46.8 & 43.1 & 50.6 & 668M & 45.5 & 45.6 & 41.6 & 49.7 & 748M & 43.0 \\
    \rowcolor{SkyBlue}
    {\musQuarter} LiVOS
    & RN-50 & RN-18 & \xmark
    & 50.6 & 46.5 & 54.7 
    & \textbf{503M} & 47.3 
    & 44.6 & 41.2 & 47.9 
    & \textbf{575M} & 45.2 \\
    \rowcolor{SkyBlue}
    {\musQuarter} LiVOS$^\ast$     
    & RN-50 & RN-18 & \xmark
    & \textbf{51.2} & 46.8 & \textbf{55.6} 
    & \textbf{503M} & 47.3 
    & \textbf{50.9} & \textbf{47.0} & \textbf{54.7} 
    & \textbf{575M} & 45.2 \\
    \rowcolor{SkyBlue}
    {\musHalf} LiVOS
    & RN-50 & RN-18 &  \xmark
    & \textbf{51.2} & \textbf{47.3} & 55.1
    & \textbf{503M} & 47.3
    & 47.0 & 44.0 & 50.0 
    & \textbf{575M} & 45.2 \\
    \bottomrule
\end{tabular}
\caption{\emph{Quantitative comparisons on long-term videos (480p).} We present the same model variants and notations as in Tab.~\ref{tab:quantitative_short_videos}. We additionally report FPS and the maximum GPU memory usage. $^\ddag$RDE consumes unexpectedly large GPU memory, likely due to implementation issues in its released code. LiVOS consistently achieves the best performance across non-STM methods on both LVOS validation and test sets while using less GPU memory.}
\label{tab:quantitative_long_videos}
\end{table*}
\begin{table*}
\centering
\resizebox{\textwidth}{!}{
\footnotesize
\begin{tabular}{l@{\hspace{1pt}}c c c c c c c c c c c c c}
    \toprule
    \multirow{2}{*}{\textbf{Method}}
    & \multirow{2}{*}{\textbf{with}} 
    & \multicolumn{3}{c}{\textbf{1024P}} 
    & \multicolumn{3}{c}{\textbf{2048P}} 
    & \multicolumn{3}{c}{\textbf{3072P}} 
    & \multicolumn{3}{c}{\textbf{4096P}} \\
    \cmidrule(lr){3-5}
    \cmidrule(lr){6-8}
    \cmidrule(lr){9-11}
    \cmidrule(lr){12-14}    
    & \textbf{STM}
    & Mem$\downarrow$ & FPS$\uparrow$ & $\mathcal{J\&F}$
    & Mem$\downarrow$ & FPS$\uparrow$ & $\mathcal{J\&F}$
    & Mem$\downarrow$ & FPS$\uparrow$ & $\mathcal{J\&F}$
    & Mem$\downarrow$ & FPS$\uparrow$ & $\mathcal{J\&F}$ \\
    \midrule
    {\musQuarter} STCN~\cite{cheng2021rethinking} \tiny{NeurIPS'21}
    & \cmark
    & 22.6G & 3.7 & 83.1 
    & \textcolor{Red}{\emph{OOM}}
    & \textcolor{Red}{\emph{N/A}} & \textcolor{Red}{\emph{N/A}}
    & \textcolor{Red}{\emph{OOM}}
    & \textcolor{Red}{\emph{N/A}} & \textcolor{Red}{\emph{N/A}}
    & \textcolor{Red}{\emph{OOM}}
    & \textcolor{Red}{\emph{N/A}} & \textcolor{Red}{\emph{N/A}} \\
    {\musQuarter} AOT~\cite{yang2021associating} \tiny{NeurIPS'21}
    & \cmark
    & \textcolor{Red}{\emph{OOM}}
    & \textcolor{Red}{\emph{N/A}} & \textcolor{Red}{\emph{N/A}}
    & \textcolor{Red}{\emph{OOM}}
    & \textcolor{Red}{\emph{N/A}} & \textcolor{Red}{\emph{N/A}}
    & \textcolor{Red}{\emph{OOM}}
    & \textcolor{Red}{\emph{N/A}} & \textcolor{Red}{\emph{N/A}}
    & \textcolor{Red}{\emph{OOM}}
    & \textcolor{Red}{\emph{N/A}} & \textcolor{Red}{\emph{N/A}} \\
    {\musQuarter} XMem~\cite{cheng2022xmem} \tiny{ECCV'22}
    & \cmark
    & 12.5G & 7.3 & 86.8 
    & \textcolor{Red}{\emph{OOM}}
    & \textcolor{Red}{\emph{N/A}} & \textcolor{Red}{\emph{N/A}}
    & \textcolor{Red}{\emph{OOM}}
    & \textcolor{Red}{\emph{N/A}} & \textcolor{Red}{\emph{N/A}}
    & \textcolor{Red}{\emph{OOM}}
    & \textcolor{Red}{\emph{N/A}} & \textcolor{Red}{\emph{N/A}} \\
    {\musQuarter} DeAOT~\cite{yang2022decoupling} \tiny{NeurIPS'22}
    & \cmark
    & 13.0G & 4.5 & 86.9 
    & \textcolor{Red}{\emph{OOM}}
    & \textcolor{Red}{\emph{N/A}} & \textcolor{Red}{\emph{N/A}}
    & \textcolor{Red}{\emph{OOM}}
    & \textcolor{Red}{\emph{N/A}} & \textcolor{Red}{\emph{N/A}}
    & \textcolor{Red}{\emph{OOM}}
    & \textcolor{Red}{\emph{N/A}} & \textcolor{Red}{\emph{N/A}} \\
    {\musQuarter} DEVA~\cite{cheng2023tracking} \tiny{ICCV'23}
    & \cmark
    & 15.7G & 5.1 & 90.0 
    & \textcolor{Red}{\emph{OOM}}
    & \textcolor{Red}{\emph{N/A}} & \textcolor{Red}{\emph{N/A}}
    & \textcolor{Red}{\emph{OOM}}
    & \textcolor{Red}{\emph{N/A}} & \textcolor{Red}{\emph{N/A}}
    & \textcolor{Red}{\emph{OOM}}
    & \textcolor{Red}{\emph{N/A}} & \textcolor{Red}{\emph{N/A}} \\
    {\musQuarter} Cutie-small~\cite{cheng2024putting} \tiny{CVPR'24}
    & \cmark
    & 9.7G & 7.6 & 89.3 
    & \textcolor{Red}{\emph{OOM}}
    & \textcolor{Red}{\emph{N/A}} & \textcolor{Red}{\emph{N/A}}
    & \textcolor{Red}{\emph{OOM}}
    & \textcolor{Red}{\emph{N/A}} & \textcolor{Red}{\emph{N/A}}
    & \textcolor{Red}{\emph{OOM}}
    & \textcolor{Red}{\emph{N/A}} & \textcolor{Red}{\emph{N/A}} \\
    {\musQuarter} Cutie-base~\cite{cheng2024putting} \tiny{CVPR'24}
    & \cmark
    & 9.9G & 6.6 & 89.5 
    & \textcolor{Red}{\emph{$150$G}}$^\ddag$ 
    & \textcolor{Red}{\emph{N/A}} & \textcolor{Red}{\emph{N/A}}
    & \textcolor{Red}{\emph{$735$G}}$^\ddag$
    & \textcolor{Red}{\emph{N/A}} & \textcolor{Red}{\emph{N/A}}
    & \textcolor{Red}{\emph{$2815$G}}$^\ddag$
    & \textcolor{Red}{\emph{N/A}} & \textcolor{Red}{\emph{N/A}} \\
    \midrule
    {\musQuarter} RDE~\cite{li2022recurrent} \tiny{CVPR'22}
    & \xmark
    & 14.4G & 10.6 & 76.3 
    & \textcolor{Red}{\emph{OOM}}
    & \textcolor{Red}{\emph{N/A}} & \textcolor{Red}{\emph{N/A}}
    & \textcolor{Red}{\emph{OOM}}
    & \textcolor{Red}{\emph{N/A}} & \textcolor{Red}{\emph{N/A}}
    & \textcolor{Red}{\emph{OOM}}
    & \textcolor{Red}{\emph{N/A}} & \textcolor{Red}{\emph{N/A}} \\
    {\musQuarter} Cutie-small$^\dag$~\cite{cheng2024putting} \tiny{CVPR'24}
    & \xmark
    & 2.3G & 11.9 & 83.7 
    & 30.3G & 2.1 & 79.8
    & \textcolor{Red}{\emph{OOM}}
    & \textcolor{Red}{\emph{N/A}} & \textcolor{Red}{\emph{N/A}}
    & \textcolor{Red}{\emph{OOM}}
    & \textcolor{Red}{\emph{N/A}} & \textcolor{Red}{\emph{N/A}} \\
    {\musQuarter} Cutie-base$^\dag$~\cite{cheng2024putting} \tiny{CVPR'24}
    & \xmark
    & 2.6G & 10.2 & \textbf{86.0} 
    & 31.1G & 1.9 & 79.6 
    & \textcolor{Red}{\emph{147G}}$^\ddag$
    & \textcolor{Red}{\emph{N/A}} & \textcolor{Red}{\emph{N/A}}
    & \textcolor{Red}{\emph{563G}}$^\ddag$
    & \textcolor{Red}{\emph{N/A}} & \textcolor{Red}{\emph{N/A}} \\
    \rowcolor{SkyBlue}
    {\musQuarter} LiVOS
    & \xmark
    & \textbf{2.0G} & \textbf{12.8} & 85.0 
    & \textbf{7.7G} & \textbf{3.5} & \textbf{80.0}
    & \textbf{17.2G} & \textbf{1.5} & \textbf{73.4}
    & \textbf{30.4G} & \textbf{0.8} & \textbf{61.5} \\
    \bottomrule
\end{tabular}}
\caption{\emph{Quantitative comparisons on high-resolution videos ($\geq$1024p).} These videos are upsampled from DAVIS 2017 validation set. All models were tested on an NVIDIA A6000 GPU. \textcolor{Red}{\emph{OOM}} indicates out-of-memory, exceeding the A6000's 48G limit. $^\ddag$Estimated memory usage. {\musQuarter} denotes models trained on YouTubeVOS and DAVIS.}
\label{tab:quantitative_highres_videos}
\end{table*}

\subsection{Comparisons}

We compare our method with baselines on three types of video benchmarks: 1) short videos, 2) long-term videos, and 3) high-resolution videos. 

\noindent\textbf{Short Videos.} Tab.~\ref{tab:quantitative_short_videos} shows results on standard VOS benchmarks: MOSE~\cite{ding2023mose}, DAVIS~\cite{pont20172017}, and YouTube VOS~\cite{xu2018youtube}. Our method achieves a score of 64.8 $\mathcal{J\&F}$ on MOSE~\cite{ding2023mose}, trailing the state-of-the-art method Cutie~\cite{cheng2024putting} by just 3.5 $\mathcal{J\&F}$. Note that MOSE is very challenging because of the complexity of multi-object interactions and dynamic environments. By default, Cutie~\cite{cheng2024putting} uses five memory frames. When limited to one memory frame, in a variant labeled Cutie$^\dag$, its performance drops significantly, falling below ours. Moreover, our method consistently outperforms other non-STM methods, including MobileVOS~\cite{miles2023mobilevos} and RDE~\cite{li2022recurrent}, across most benchmarks.

\noindent\textbf{Long-term Videos.} Tab.~\ref{tab:quantitative_long_videos} shows results on the long video benchmark LVOS~\cite{hong2022lvos}. Compared with the state-of-the-art recurrent memory network RDE~\cite{li2022recurrent}, our method is better across all benchmarks. Compared with state-of-the-art STM-based methods, our method delivers competitive performance while being faster and more memory efficient. Even with Cutie's memory bank limited to one memory frame, our method requires less memory while achieving better segmentation quality. Note that memory networks such as Cutie and DEVA employ a bounded memory bank (by default, the number is five), so they have no memory issues when video length increases. To highlight our memory efficiency, we also compare our method with baselines on high-resolution videos, introduced below.

\noindent\textbf{High-resolution Videos.} Tab.~\ref{tab:quantitative_highres_videos} shows the comparisons on high-resolution videos. Note that these high-resolution videos are upsampled from the DAVIS 2017 validation dataset. All state-of-the-art space-time memory networks encounter out-of-memory issues at 2048p. For example, Cutie-base requires approximately 150GB of memory to run inference at 2048p, far exceeding the capabilities of consumer-grade GPUs. In contrast, our method requires only 7.7GB for 2048p inference and is the only one capable of inference at 4096p. While not specifically optimized for high-resolution videos, our method sets the stage for developing foundation models suited to high-resolution regimes.

\subsection{Ablations}
Tab.~\ref{tab:ablation_livos} shows the results of an ablation study to validate our design choices. The models are trained on YouTube-VOS and DAVIS and evaluated on the DAVIS validation set. \emph{No gate} is a model variant that disables the gate mechanism in linear matching. \emph{No sensory memory} is a model variant that disables the sensory memory. \emph{No object memory} is a model variant that disables the object memory. The gate mechanism in linear matching improves 1.4\% $\mathcal{J\&F}$, with only a minor trade-off in speed and memory usage. The integration of sensory and object memories substantially boost performance, though it comes at the cost of reduced speed and higher memory consumption. In this ablation, we retain linear matching instead of swapping it with softmax matching, as that would make our method similar to Cutie~\cite{cheng2024putting}.

\begin{table}
    \centering
    \resizebox{0.47\textwidth}{!}{    
    \begin{tabular}[c]{l c c c c c}
         \toprule
         \textbf{Method}
         & $\mathcal{J\&F}$ & $\mathcal{J}$ & $\mathcal{F}$ 
         & FPS $\uparrow$ & Mem $\downarrow$ \\
         \midrule
         No gate            
         & 83.0 & 80.0 & 86.0 
         & 40.8 & 573M \\
         No object memory
         & 80.4 & 77.4 & 83.4 
         & \textbf{70.5} & 565M \\
         No sensory memory
         & 79.1 & 76.3 & 81.9 
         & 52.6 & \textbf{538M} \\
         \midrule
         Full
         & \textbf{84.4} & \textbf{81.2}  & \textbf{87.6} 
         & 40.3  & 574M \\
         \bottomrule
    \end{tabular}}
    \caption{Ablation study for LiVOS on DAVIS 2017 val~\cite{perazzi2016benchmark}.}
    \label{tab:ablation_livos}
\end{table}
\section{Limitations}
\label{sec:limitations}

Our method uses a single recurrent state for the challenging VOS task, which works well for short, low-resolution videos but may lead to suboptimal results on longer, high-resolution ones. We believe a more advanced state~\cite{gu2023mamba}, as well as multi-scale linear attention~\cite{cai2022efficientvit}, could improve performance in these cases. We leave these extensions for future work.
\section{Conclusion}
\label{sec:conclusion}

We proposed the first light memory network that employs gated linear matching, instead of softmax matching, for memory efficient video object segmentation. Our method reformulates the memory matching process into a recurrent framework where the large, quadratic attention matrix is reduced to a small, constant-size recurrent state. Our evaluation results on long and high-resolution videos demonstrated the effectiveness and efficiency of our method, opening the door for high-resolution video foundation models.

\footnotesize{\noindent\textbf{Acknowledgement.} This research was, in part, funded by the National Institutes of Health (NIH) under other transactions 1OT2OD038045-01 and NIAMS 1R01AR082684. The views and conclusions contained in this document are those of the authors and should not be interpreted as representing official policies, either expressed or implied, of the NIH.}
{
    \small
    \bibliographystyle{ieeenat_fullname}
    \bibliography{main}
}


\end{document}